\documentclass[conference]{IEEEtran}
\IEEEoverridecommandlockouts
\usepackage{amsmath,amssymb,amsfonts}
\usepackage{algorithmic}
\usepackage{graphicx}
\usepackage{xcolor}
\usepackage{url}
\usepackage{hyperref}
\usepackage{lineno}

\usepackage{upquote} 
\usepackage{textcomp} 
\usepackage{booktabs}
\usepackage{adjustbox}
\usepackage{comment}

\begin{document}

\def\BibTeX{{\rm B\kern-.05em{\sc i\kern-.025em b}\kern-.08emT\kern-.1667em\lower.7ex\hbox{E}\kern-.125emX}}

\title{Everyday Speech in the Indian Subcontinent }






\author{
\IEEEauthorblockN{ Utkarsh Pathak}
\IEEEauthorblockA{\textit{Dept of Computer Science \& Engg.} \\
\textit{Indian Institute of Technology Madras}\\
Chennai, India \\
utkarshpathak16@gmail.com}
\and
\IEEEauthorblockN{ Chandra sai krishna G}
\IEEEauthorblockA{\textit{Dept. of Computer Science \& Engg.} \\
\textit{Indian Institute of Technology Madras}\\
Chennai, India  \\
saikrishnag432@gmail.com}
\and
\IEEEauthorblockN{ Sujitha Sathiyamoorthy}
\IEEEauthorblockA{\textit{Dept. of Computer Science \& Engg.} \\
\textit{Indian Institute of Technology Madras}\\
Chennai, India \\
sujivk16@gmail.com}
\and
\IEEEauthorblockN{Keshav Agarwal}
\IEEEauthorblockA{\textit{Dept. of Computer Science \& Engg.} \\
\textit{Indian Institute of Technology Madras}\\
Chennai, India \\
kshvgrwal@gmail.com}
\and
\IEEEauthorblockN{Hema A Murthy}
\IEEEauthorblockA{\textit{Dept. of Computer Science \& Engg.} \\
\textit{Indian Institute of Technology Madras and Shiv Nadar University}\\
Chennai, India \\
hema@cse.iitm.ac.in}

}

\maketitle
\begin{abstract}

India has 1369 languages of which 22 are official. About 13 different scripts are used to represent these languages. A Common Label Set(CLS) was developed based on phonetics to address the issue of large vocabulary of units required in the End-to-End(E2E) framework for multilingual synthesis. The Indian language text is first converted to CLS. This approach enables seamless code switching across 13 Indian languages and English in a given native speaker's voice, which corresponds to everyday speech in the Indian subcontinent, where the population is multilingual. 
\end{abstract}

\section{Introduction}
\label{sec:intro}
The Indian subcontinent is home to diverse cultures and languages. 
While the European union has 3 to 4 scripts, Indian languages have around 66 scripts, of which 13 are used in the 22 official languages. 
The Indian subcontinent is also considered a linguistic area, or ``sprachbund" \cite{Trubetzkoy}, where extended contact between languages has led to convergence in phonological, grammatical, and other features.  Similarities in prosodic patterns, such as stress and intonation, enhance the perceived resemblance among Indian languages. Indo-Aryan languages form a dialect-continuum i.e., languages geographically close to each other are remarkably similar but differences increase over distance.\\ Most Indians are multilingual and can speak 2-3 languages (Hindi, English and some regional languages)\cite{census2011}. Owing to multilinguality, code mixing and code switching is very natural in ``everyday speech" in the Indian subcontinent\cite{thomas18_interspeech}. 
Despite advancements in TTS models \cite{oord2016wavenet, prenger2019waveglow,kim2020glow,popov2021grad,kim2021conditional}, building TTS for low resource languages, while at the same time ensuring code switching/mixing is still challenging. Recent advances in multilingual synthesis \cite{google_100_multi,lux2024meta,yourtts2022,valle-X} have introduced various frameworks aimed at generating multilingual, multispeaker audio with large amounts of data.  
For low resource scenario, most systems involve fine-tuning a pretrained data model with small amount of data from a target, or cross-lingual transfer learning.  

Various attempts at developing a unified linguistic front-end for all languages have been attempted in \cite{gutkin2017uniform,anushalangaugeindependent,Prakash2023734, prakash19_ssw, prakash2020generic}.  Closely related languages are grouped and TTS models are trained.  Zero or small amount adaptation data is used to produce speech in a target language.  In particular, work done by  \cite{gutkin2017uniform}, shows that a Mean Opinion score (MOS) of 2.5 is achievable for zero resource languages.

In \cite{thomas18_interspeech}, Code-switching was feasible especially for the languages included in the training data. Additionally, the speaker was selected based on the language identified from the first character of the text.






In this paper, we study a different problem, where the  native speaker characteristics are preserved while speaking a nonnative, or/and code mixing/switching language(typical in  Indian language conversations  \cite{kurian2011indian}).    This is carried out by  modifying the CLS \cite{ramani2013common} to first create a super set of labels  to accommodate all the 22 official languages.   Phonetic, and linguistic characteristics of languages are studied, and the same is incorporated in the synthesiser.  Zero shot speech synthesisers are built for Sanskrit, and two dialects of Konkani, using a modified CLS, a modified parser, and appropriate synthesiser.  In addition, seamless  code mixed/code-switched speech synthesis is performed.

The rest of the paper is organised as follows.
Section \ref{sec:motivation} gives an overview and motivates the work done. Section \ref{sec:cls} reviews the work done on the common label set for Indian languages and the modifications made to the same to support code mixed/code switched synthesis.  In Section \ref{sec:languages}, we discuss the properties of Indian languages in terms of similarities in scripts and phonotactics for Sanskrit and Konkani. Section \ref{sec:cd} discuss the evaluation of synthesisers for zero shot  speech and code mixed speech, respectively.  Finally, Section \ref{sec:conclusions} discusses the way forward given the new insights obtained owing to the experiments reported in this paper.

\section{Motivation}
\label{sec:motivation}   
Owing to several factors, namely geographical, social, cultural and migration, India has a rich linguistic diversity.  There are about 19569 dialects according to the 2011 census \cite{census_2011}, and most Indians are multilingual.  Code mixing (English and native tongue) has become very common in the rural areas due to digitalisation.  Migration of people for work, further increases the code switching.
The article \cite{accentarticle} shows that native accented speech has high intelligibility with less listening effort, compared to non-native accented speech/languages.  Visually challenged persons using screen readers preferred the native accented English over the British/US English variant of the TTS \cite{kurian2011indian}. The study \cite{accentbias} on Tutorbots also shows that native accented bots were more competent, likeable, knowledgeable and helpful.  These efforts and multilinguality make a strong case for building native and seamless code mixing/switching synthesisers.





\section{Importance of Common Label Set (CLS) in code mixed and code switched  speech synthesis in Indian languages}
\label{sec:cls}



\subsubsection{{\it Properties of Konkani and Sanskrit}}
\label{sec:languages}
Konkani is considered an Indo-Aryan language. Historically, Konkani evolved from old Indo Aryan through Maharashtri Prakrit and Middle Indo Aryan languages. It has retained archaic features compared to Marathi and displays a strong inflectional structure similar to Sanskrit. Its closest linguistic relative within this group is Marathi. High Nasalisation makes it unique among Indo-Aryan languages \cite{fadte2023empirical}. Konkani is written in multiple scripts including Devanagari, Roman, Kannada, Malayalam, and Perso-Arabic. Different communities and regions use distinct scripts; for instance, Goan Hindus use Devnagari while Goan Christians use the Roman script.\footnote{\url{http://lisindia.ciil.org/Konkani/Konkani_hist.html}} There are various dialects of Konkani; for instance Marathi/Goan konkani in Northern Regions while Canara konkani in Southern regions. 


Sanskrit is a classical language, and many Indian languages have their origins in Sanskrit.
Telugu on the other hand has its origins in proto-dravidian. Telugu literature is replete with Sanskrit words. Telugu shares similarity with Sanskrit in that majority of the vocabulary, characteristics, structure, grammar rules are very similar  \cite{rao2014comparative}. Similarly Kannada (Dravidian origin) has also been influenced by Sanskrit\cite{kannadasanskrit}.

\subsection{Updation to Common Label Set (CLS) table: Mapping of  missing sounds in the CLS for Code mixing and Code switched synthesis } 

We demonstrate a monolingual synthesiser to synthesise code-mixed/code switched text.

The datasets used in this work, sourced from \cite{baby2016resources} and available here\footnote{\url{https://www.iitm.ac.in/donlab/indictts/database}}, provide phonetically rich, high-quality recordings for over 13 Indian languages, collected from native professional voice talents in noise-free studio environments at 48kHz, 16-bit resolution. 
\footnote{Sanskrit and Konkani datasets have been recently added to this corpus.}

Three different types of tests were performed, namely, MOS, AXY discrimination and language identification test.

\section{Evaluation of Code mixed and Code switched synthesis }
\label{sec:cd}
The unified parser generates CLS labels for all languages.   These CLS labels are agnostic to the choice of the synthesiser.   In this task, we choose a particular voice, and synthesise nonnative language/code switched text using the particular voice.  The text is first converted to the CLS using modifications presented in Section \ref{sec:cls} for code switching.  

\subsection {Intelligibility and naturalness: MOS}
In this test, Hindi and Kannada (nonnative) synthesisers  were used to synthesise the text in 5 different languages.   Native speakers of these 5 languages were asked to evaluate the quality of the output using standard MOS for naturalness and intelligibility (scale 1-5, 1-poor, 5-excellent).  From Table \ref{tab:cdmos}, it is clear that monolingual synthesis using a voice other than that of the native tongue produces reasonable quality.

\subsection{AXY discrimination test}


The AXY test is a perceptual evaluation method Subjects are presented with 3 audio clips: a fixed reference (A), and two competing samples (X,Y), both played in random order, and  subjects evaluate the similarity of A to X, and Y \cite{pmlr-ryan18a}.  This test was used to evaluate the synthesis quality of  two dialects of Konkani (North Canara and South Canara).  Subjects were speakers of the corresponding native dialect of Konkani.  As conjectured, Marathi TTS for North Canara Konkani was preferred over Kannada TTS by North Canara Konkani subjects and vice-versa.

Clearly, choosing the appropriate TTS model for targeted dialects/unseen/low resource language matters.  Nevertheless, from the MOS scores, AXY test, it is clear that text given in a particular language could be synthesised using a voice from a different language provided the language shares properties with the TTS system used \cite{masica1993indo, krishnamurti2003dravidian}. 
\begin{table}[htbp]
{\tiny
\caption{Mean opinion score (MOS) of Hindi and Kannada synthesisers for different languages.}
\centering
\resizebox{\columnwidth}{!}{%
\begin{tabular}{cccccc}
\hline
\textbf{\begin{tabular}[c]{@{}c@{}}Text \\ Language \\(evaluators)\end{tabular}} & \textbf{\begin{tabular}[c]{@{}c@{}}Hi \\ intelligibility \\ (MOS)\end{tabular}} & \textbf{\begin{tabular}[c]{@{}c@{}}Hi \\ naturalness\\ (MOS)\end{tabular}} & \textbf{\begin{tabular}[c]{@{}c@{}}Kan \\ intelligibility\\ (MOS)\end{tabular}} & \textbf{\begin{tabular}[c]{@{}c@{}}Kan \\ naturalness\\ (MOS)\end{tabular}} \\ \hline
\textbf{Telugu (21)}                                              & 4.13                                                                          & 3.19                                                                          & 4.22                                                                          & 3.85                                                                          \\ 
\textbf{Hindi (22)}                                               & 4.50                                                                          & 4.46                                                                          & 4.26                                                                          & 4.02                                                                          \\ 
\textbf{Marathi (14)}                                              & 3.90                                                                          & 3.81                                                                          & 4.02                                                                          & 4.00                                                                          \\ 
\textbf{Kannada (12)}                                              & 2.67                                                                          & 2.03                                                                          & 4.30                                                                          & 4.35                                                                          \\ 
\textbf{Odia (11)}                                                & 3.58                                                                          & 2.82                                                                          & 3.60                                                                          & 2.21                                                                          \\ \hline
\textbf{\begin{tabular}[c]{@{}c@{}}Mean  (MOS)\end{tabular}}     & 3.75                                                                          & 3.26                                                                          & 4.08                                                                          & 3.68                                                                          \\ \hline
\end{tabular}%
\label{tab:cdmos}
}}
\end{table}

\section{Conclusions}
\label{sec:conclusions}

In this paper, the objective is multifold.  The common label set and parser are modified to accommodate sounds corresponding to unseen languages and code switched text. Code mixing and code switching is very common in everyday communication in the Indian subcontinent owing to the fact that most Indians speak at least 2-3 languages.  Seamless code mixed and code switched text is synthesised without an increase in the footprint of the synthesis system in a language of one's choice.  



\bibliographystyle{IEEEtran}
\bibliography{mybib}
\end{document}